\documentclass[10pt,twocolumn,letterpaper]{article}

\usepackage{iccv}
\usepackage{times}
\usepackage{epsfig}
\usepackage{graphicx}
\usepackage{amsmath}
\usepackage{amssymb}

\usepackage[breaklinks=true,bookmarks=false]{hyperref}

\iccvfinalcopy 

\def\iccvPaperID{****} 

\ificcvfinal\fi

\begin{document}

\title{Action Unit Detection with Joint Adaptive Attention and Graph Relation}

\author{Chenggong Zhang, Juan Song, Qingyang Zhang, Weilong Dong, Ruomeng Ding, Zhilei Liu\thanks{Corresponding Author}\\
College of Intelligence and Computing, Tianjin University, Tianjin, China, 300350\\
{\tt\small \{zhangchenggong, songjuan2\_1, willowd, ruomengding, zhileiliu\}@tju.edu.cn}
}

\maketitle
\ificcvfinal\thispagestyle{empty}\fi

\begin{abstract}
   This paper describes an approach to the facial action unit (AU) detection. In this work, we present our submission to the Field Affective Behavior Analysis (ABAW) 2021 competition. The proposed method uses the pre-trained JAA model as the feature extractor, and extracts global features, face alignment features and AU local features on the basis of multi-scale features. We take the AU local features as the input of the graph convolution to further consider the correlation between AU, and finally use the fused features to classify AU. The detected accuracy was evaluated by 0.5×accuracy + 0.5×F1. Our model achieves 0.674 on the challenging Aff-Wild2 database.
\end{abstract}
\section{Introduction}

Automatic Action Units (AUs) recognition is useful and important in facial expression analysis~\cite{zhi2020comprehensive}. Facial Action Coding System (FACS)~\cite{friesen1978facial} defines a unique set of about 40 facial muscle actions known as Action Units (AUs), each AU represents a basic facial movement or expression change. Due to the complexity of facial texture structures and the fact that AU does not appear in isolation, it is important to consider the local characteristics of AU and its symbiotic relationships. In addition, existing landmarks detection methods are highly accurate, and the detected landmarks can provide accurate AU location information.

We take the pre-trained JAA-Net~\cite{shao2021jaa} as the feature extractor and train the graph convolution model on the basis of JAA to model the relationship between AU. Before feature extraction, we use Dlib to detect landmarks as the ground-truth of the JAA face-alignment module.
Feature extraction is carried out on the basis of the extracted multi-scale features, including facial Landmarks features, global features and local features combined with AU relationships. As an auxiliary task for AU detection, the face alignment task learns face features with detected landmarks as constraints. Global feature is the feature extraction of the whole face. Local feature learning is the refinement of each AU attention map in order to learn local features specific to AU. On the basis of the local features of AU, the graph convolution model is introduced, and the local features of AU are taken as the nodes of the graph convolution to learn the relations among various AU. 
In the AU classification, all the learned features are fused and then taken to judge whether the target AUs are observed.

\section{Proposed Method}

Our method consists of two steps in training phase. In the training phase, first, we used the pre-trained JAA model to extract features, obtained refined AU attention features. Secondly, we have added a graph convolution model on the basis of JAA, and use the local features of each AU output by JAA as a node to learn the correlation between AUs. Finally, the face alignment feature, global feature and AU relationship feature are fused to perform AU detection. The overall pipeline of our method is illustrated in Figure 1 and it will be detailed in this section.

\begin{figure*}
\begin{center}
\includegraphics[scale=0.45]{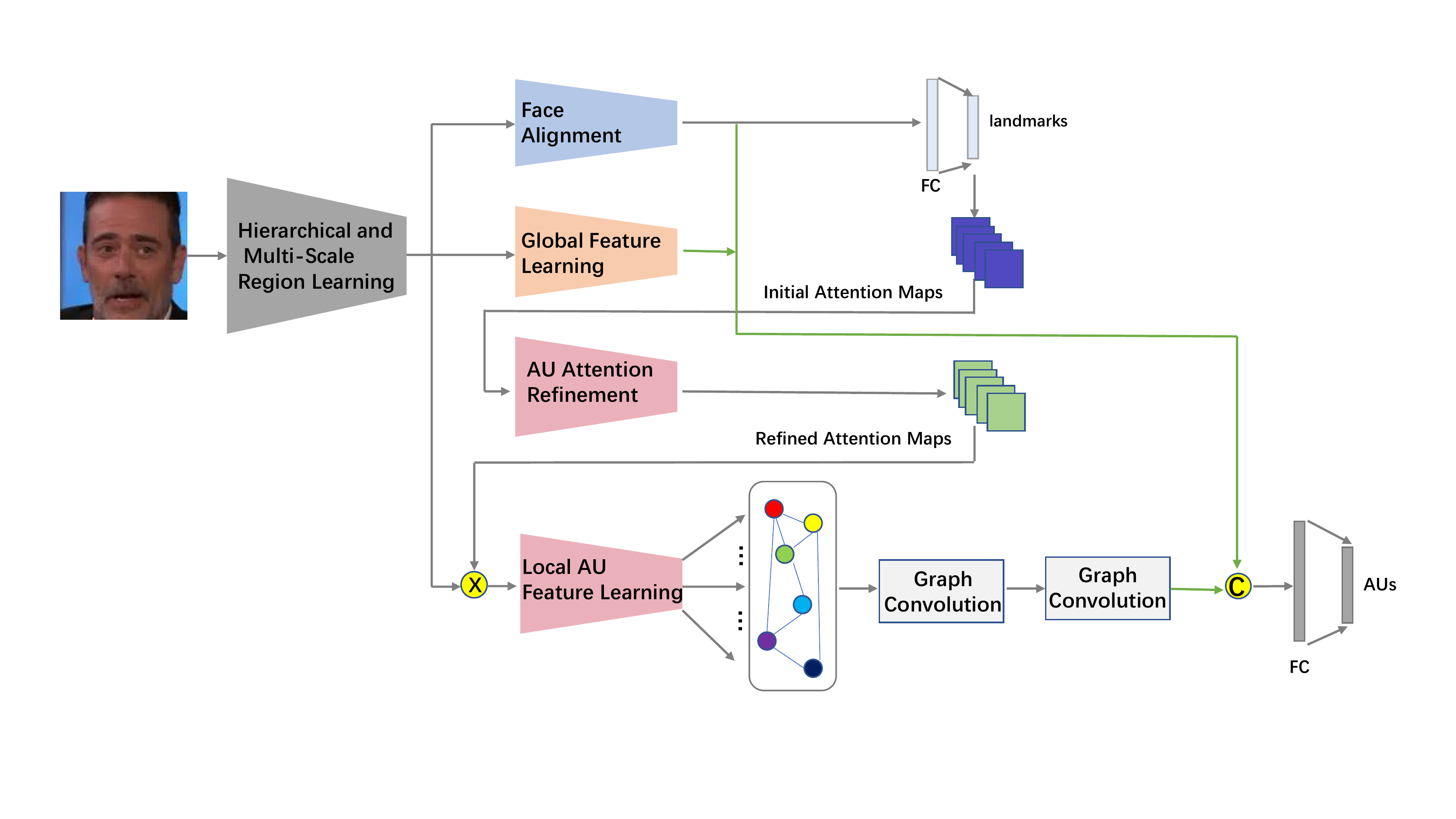}
\end{center}
   \caption{Overview of our method. Multi-scale shared feature is learned firstly. The AU local module (red) are used to learn the AU local representation, and the GCN module uses the AU local representation as input to learn the correlation between AUs.}
\label{fig:fig1}
\end{figure*}
\subsection{Pre-trained JAA model}

We input 112×112×3 color faces into JAA and use the detected landmarks as the ground-truth for face alignment. The following is the introduction of feature extraction module.
\subsubsection{Multi-Scale Region Learning and Face Alignment}

Considering AUs in different local facial regions have various structure and texture information, different local regions should be processed with different filters. Therefore, in order to deal with AU regions of different sizes and texture information of different regions, the article proposes a hierarchical and multi-scale region layer to learn AU information of different scales. It consists of an ordinary convolutional layer and another three hierarchical partitioned convolutional layers, which are more suitable for the extraction of facial features than ordinary convolutional layers.

The face alignment module outputs a face alignment feature, which contains global face shape information and local landmark information. The loss of the face alignment module is as follows:
\begin{equation}
 E_{\text {align }}=\frac{1}{2 d^{2}} \sum_{j=1}^{n_{\text {align }}}\left[\left(y_{2 j-1}-\hat{y}_{2 j-1}\right)^{2}+\left(y_{2 j}-\hat{y}_{2 j}\right)^{2}\right] 
\end{equation}
The face alignment loss is the same as that for JAA-Net in face alignment module. 

\subsubsection{Adaptive Attention Learning}

Adaptive attention learning consists of two steps: AU attention refinement and local AU feature learning. The first step is to refine the predefined attention map of each AU with a separate branch, and the second step is to learn and extract local AU feature. The attention map of each AU is accurately positioned by the landmarks coordinates learned by face alignment. The detection loss of local AU is used to supervise the generation of the attention map. We use the same loss functions as they are defined in ~\cite{shao2021jaa}. We use ${{f}_{i}}$ to represent the local feature of each AU.
\subsection{Training GCN Model}

First, the local features ${{f}_{i}}$ output in JAA are used as nodes to construct the relational matrix, which is represented by ${M}$. The number ${M_{i j}}$ represent the conditional probability of AU ${i}$ and AU ${j}$. Then, a threshold is set to convert ${M_{i j}}$ into a 0-1 matrix ${M^{bool}}$ as following:
\begin{equation}
 M_{i j}^{\text {bool}}=\left\{\begin{array}{l}1 \text { if } M_{i j}>=\text { threshold } \\ 0 \text { if } M_{i j}<\text { threshold }\end{array}\right. 
\end{equation}
if ${M_{i j}^{{bool}}}$ = 1, it shows that AU ${i}$ is strongly related to AU ${j}$. we apply the Graph Convolutional Networks(GCNs) proposed in ~\cite{kipf2016semi}. We
can represent one layer of graph convolution as:
\begin{equation}
 Z_{i+1}=G \times Z_{i} \times W_{i} 
\end{equation}
where ${G}$ represents the adjacency graph we have introduced above with R×R dimension. ${Z_{i}}$ denotes the input features in the ${i}$-th graph convolution operation, and ${W_{i}}$ is the weight matrix of the layer. The updated features after convolution of the graph are flattened. Then, the flat features are forwarded to the Fully Connected Network (FCN) for AU detection.
\subsection{Loss Functions}
Facial AU detection can be regarded as a multi-label binary classification problem, we represent the weighted multi-label softmax loss as ${L_{{softmax}}(Y, \hat{Y})}$.
Where ${\hat{Y}}$ denotes the predicted occurrence probability, ${Y}$ denotes the ground-truth occurrence probability.

The weight is to alleviate the data imbalance problem. Since AUs are not mutually independent, imbalanced training data has a bad influence on this multi-label learning task. We set ${w_{i}=\frac{\left(1 / r_{i}\right) C}{\sum_{i=1}^{C}\left(1 / r_{i}\right)}}$ as the training weight. Where, ${r_i}$ is the occurrence rate of the ${i}$-th AU in the training set. We further introduced the weighted multi-label Dice coefficient loss ${L_{dice}}$~\cite{milletari2016fully} to prevent the loss of some AU in training samples, as shown below:
\begin{equation}
 L_{\text {dice }}(Y, \hat{Y})=\frac{1}{C} \sum_{i=1}^{C} w_{i}\left(1-\frac{2 Y_{i} \hat{Y}_{i}+\epsilon}{Y_{i}^{2}+\hat{Y}_{i}^{2}+\epsilon}\right) 
\end{equation}
where ${\epsilon}$ is the smooth term. Finally, the AU detection loss is defined as:
\begin{equation}
 L_{a u}=L_{\text {softmax }}+\lambda_{2} L_{\text {dice }} 
\end{equation}
where ${\lambda_{2}}$ is a trade-off parameter.
\section{Experiment}

\subsection{Datasets and Metrics}
We evaluate the proposed method on the Aff-Wild2 dataset~\cite{kollias2021distribution}~\cite{kollias2021affect}~\cite{kollias2019expression}~\cite{kollias2019face}~\cite{kollias2019deep}~\cite{zafeiriou2017aff}, which was used in the 2nd Workshop and Competition on Affective Behavior Analysis in-the-wild (ABAW)~\cite{2106.15318}~\cite{kollias2020analysing}. Aff-Wild2 is currently the largest (and audiovisual) in the-wild database annotated in terms of the seven basic expressions and twelve action units. For both the Aff-Wild2 training and validation set we use the provided cropped and aligned face images.These images have dimensions 112×112×3. We use Dlib toolkit to detect 49 landmarks as ground-truth for training JAA-Net.
In the competition, the perfromance metric of 12 Action Unit Detection is defined as:
\begin{equation}
\label{eq_fupd}
0.5* F1Score + 0.5* Accuracy
\end{equation}
where F1 Score is the unweighted mean and Accuracy is the total accuracy.

\subsection{Implementation Details}
For the first stage, we follow the experiment setting in~\cite{shao2021jaa} to pre-train JAA-Net. We train JAA-Net for 4 epochs with initial learning rate of 0.01, in which the learning rate is multiplied by a factor of 0.3 at every 2 epochs. For the second stage, we fix the hierarchical and multi-scale region learning module, face alignment module, global feature learning module, and adaptive attention learning module, and then embed the GCN module into the network. In the second stage, we only train the GCN module and full connection layers for 3 epochs without the face alignment loss. In both two stages, we use the stochastic gradient descent (SGD) solver, a Nesterov momentum~\cite{sutskever2013importance} of 0.9, and a weight decay of 0.0005. Our experiments are implemented on Pytorch~\cite{paszke2017automatic} with NVIDIA TITAN V GPUs.

\subsection{Results}
Table 1 presents results of baseline and our method on
validation dataset. The baseline result is in~\cite{2106.15318}. 
\begin{table}
\begin{center}
\begin{tabular}{|c|c|c|c|}
\hline
  & F1 score& Accuracy& Competition Metric \\
\hline\hline
Baseline & 0.400   & 0.220   & 0.310 \\
JAA-Net  & 0.444   & 0.879   & 0.662 \\
Ours     & 0.467   & 0.880   & 0.674\\
\hline
\end{tabular}
\end{center}
\caption{Results on validation dataset}
\end{table}

As shown in Table 1, our method outperforms the baseline and JAA-Net. This result indicates that GCN can capture the strong relationship between AUs, and is useful for AU detection.

\section{Conclusions}
In this paper we present an approach based on JAA-Net and Graph Convolutiona Network to detect Action Units. Firstly, we pre-traine JAA-Net as feature extractor to extract local AU features. Second, GCN is embeded into the network to learn relationship between AUs. We compared our method and a baseline in the competition evaluation metric, and the result presented our method outperforms the baseline.

{\small
\bibliographystyle{ieee_fullname}
\bibliography{JAA-GCN}

\begin{thebibliography}{10}\itemsep=-1pt

\bibitem{friesen1978facial}
E Friesen and Paul Ekman.
\newblock Facial action coding system: a technique for the measurement of
  facial movement.
\newblock {\em Palo Alto}, 3(2):5, 1978.

\bibitem{kipf2016semi}
Thomas~N Kipf and Max Welling.
\newblock Semi-supervised classification with graph convolutional networks.
\newblock {\em arXiv preprint arXiv:1609.02907}, 2016.

\bibitem{2106.15318}
Dimitrios Kollias, Irene Kotsia, Elnar Hajiyev, and Stefanos Zafeiriou.
\newblock Analysing affective behavior in the second abaw2 competition, 2021.

\bibitem{kollias2020analysing}
D Kollias, A Schulc, E Hajiyev, and S Zafeiriou.
\newblock Analysing affective behavior in the first abaw 2020 competition.
\newblock In {\em 2020 15th IEEE International Conference on Automatic Face and
  Gesture Recognition (FG 2020)(FG)}, pages 794--800.

\bibitem{kollias2019face}
Dimitrios Kollias, Viktoriia Sharmanska, and Stefanos Zafeiriou.
\newblock Face behavior a la carte: Expressions, affect and action units in a
  single network.
\newblock {\em arXiv preprint arXiv:1910.11111}, 2019.

\bibitem{kollias2021distribution}
Dimitrios Kollias, Viktoriia Sharmanska, and Stefanos Zafeiriou.
\newblock Distribution matching for heterogeneous multi-task learning: a
  large-scale face study.
\newblock {\em arXiv preprint arXiv:2105.03790}, 2021.

\bibitem{kollias2019deep}
Dimitrios Kollias, Panagiotis Tzirakis, Mihalis~A Nicolaou, Athanasios
  Papaioannou, Guoying Zhao, Bj{\"o}rn Schuller, Irene Kotsia, and Stefanos
  Zafeiriou.
\newblock Deep affect prediction in-the-wild: Aff-wild database and challenge,
  deep architectures, and beyond.
\newblock {\em International Journal of Computer Vision}, pages 1--23, 2019.

\bibitem{kollias2019expression}
Dimitrios Kollias and Stefanos Zafeiriou.
\newblock Expression, affect, action unit recognition: Aff-wild2, multi-task
  learning and arcface.
\newblock {\em arXiv preprint arXiv:1910.04855}, 2019.

\bibitem{kollias2021affect}
Dimitrios Kollias and Stefanos Zafeiriou.
\newblock Affect analysis in-the-wild: Valence-arousal, expressions, action
  units and a unified framework.
\newblock {\em arXiv preprint arXiv:2103.15792}, 2021.

\bibitem{milletari2016fully}
F Milletari, N Navab, SA Ahmadi, and V-net.
\newblock Fully convolutional neural networks for volumetric medical image
  segmentation.
\newblock In {\em Proceedings of the 2016 Fourth International Conference on 3D
  Vision (3DV)}, pages 565--571.

\bibitem{paszke2017automatic}
Adam Paszke, Sam Gross, Soumith Chintala, Gregory Chanan, Edward Yang, Zachary
  DeVito, Zeming Lin, Alban Desmaison, Luca Antiga, and Adam Lerer.
\newblock Automatic differentiation in pytorch.
\newblock 2017.

\bibitem{shao2021jaa}
Zhiwen Shao, Zhilei Liu, Jianfei Cai, and Lizhuang Ma.
\newblock Jaa-net: joint facial action unit detection and face alignment via
  adaptive attention.
\newblock {\em International Journal of Computer Vision}, 129(2):321--340,
  2021.

\bibitem{sutskever2013importance}
Ilya Sutskever, James Martens, George Dahl, and Geoffrey Hinton.
\newblock On the importance of initialization and momentum in deep learning.
\newblock In {\em International conference on machine learning}, pages
  1139--1147. PMLR, 2013.

\bibitem{zafeiriou2017aff}
Stefanos Zafeiriou, Dimitrios Kollias, Mihalis~A Nicolaou, Athanasios
  Papaioannou, Guoying Zhao, and Irene Kotsia.
\newblock Aff-wild: Valence and arousal ‘in-the-wild’challenge.
\newblock In {\em Computer Vision and Pattern Recognition Workshops (CVPRW),
  2017 IEEE Conference on}, pages 1980--1987. IEEE, 2017.

\bibitem{zhi2020comprehensive}
Ruicong Zhi, Mengyi Liu, and Dezheng Zhang.
\newblock A comprehensive survey on automatic facial action unit analysis.
\newblock {\em The Visual Computer}, 36(5):1067--1093, 2020.

\end{thebibliography}
}

\end{document}